\documentclass[letterpaper]{article} 
\usepackage{aaai24}  
\usepackage{times}  
\usepackage{helvet}  
\usepackage{courier}  
\usepackage[hyphens]{url}  
\usepackage{graphicx} 
\usepackage{natbib}  
\usepackage{caption} 
\usepackage{algorithm}
\usepackage{algorithmic}
\usepackage{amssymb}
\usepackage{amsmath}
\usepackage{makecell}
\usepackage {booktabs}
\usepackage{subfig}
\usepackage{multirow}
\usepackage{newfloat}
\usepackage{listings}
\DeclareCaptionStyle{ruled}{labelfont=normalfont,labelsep=colon,strut=off} 
\floatstyle{ruled}
\newfloat{listing}{tb}{lst}{}
\floatname{listing}{Listing}
\title{UniCell: Universal Cell Nucleus Classification via Prompt Learning}
\author{
    Junjia Huang\textsuperscript{\rm 1,\rm 2}\equalcontrib, Haofeng Li\textsuperscript{\rm 2}\equalcontrib, Xiang Wan\textsuperscript{\rm 2}, Guanbin Li\textsuperscript{\rm 1,\rm 3}\thanks{Guanbin Li is the corresponding author.} \\
}
\affiliations{

    \textsuperscript{\rm 1}School of Computer Science and Engineering,
    Sun Yat-sen University, Guangzhou, China \\
    \textsuperscript{\rm 2}Shenzhen Research Institute of Big Data, 
    The Chinese University of Hong Kong, Shenzhen, China\\
    \textsuperscript{\rm 3} GuangDong Province Key Laboratory of Information Security Technology\\
    huangjj77@mail2.sysu.edu.cn, \{lhaof,wanxiang\}@sribd.cn, liguanbin@mail.sysu.edu.cn
}

\usepackage{bibentry}

\begin{document}

\maketitle

\begin{abstract}
The recognition of multi-class cell nuclei can significantly facilitate the process of histopathological diagnosis. Numerous pathological datasets are currently available, but their annotations are inconsistent. Most existing methods require individual training on each dataset to deduce the relevant labels and lack the use of common knowledge across datasets, consequently restricting the quality of recognition. In this paper, we propose a universal cell nucleus classification framework (UniCell), which employs a novel prompt learning mechanism to uniformly predict the corresponding categories of pathological images from different dataset domains. In particular, our framework adopts an end-to-end architecture for nuclei detection and classification, and utilizes flexible prediction heads for adapting various datasets. Moreover, we develop a Dynamic Prompt Module (DPM) that exploits the properties of multiple datasets to enhance features. The DPM first integrates the embeddings of datasets and semantic categories, and then employs the integrated prompts to refine image representations, efficiently harvesting the shared knowledge among the related cell types and data sources. Experimental results demonstrate that the proposed method effectively achieves the state-of-the-art results on four nucleus detection and classification benchmarks. Code and models are available at https://github.com/lhaof/UniCell 
\end{abstract}

\section{Introduction}

Histopathological analysis
is widely considered as the gold standard for the diagnosis and prognosis of human cancers~\cite{wu2022recent, zeiser2021breast}. Locating and classifying the cells in histopathology images is a preliminary step in analyzing, diagnosing and grading cancerous cells. In practical applications, the procedures such as cell counting~\cite{fridman2012immune}, 
tumor grading~\cite{fleming2012colorectal} and computer-aided diagnosis (CAD)~\cite{saha2016computer} all require to identify nuclei as a fundamental task. Some studies~\cite{abousamra2021multi,huang2023affine} aim to identify both the location and type of cells, while some other works~\cite{stringer2021cellpose,lou2022pixel,lou2023multi,ma2023multi,yu2023diffusion} 
attempt to determine the nucleus boundaries. However, due to the time-consume process of pixel-wise labeling, some datasets only provide the centroid coordinates of nuclei instead of boundaries. Thus, in this study we concentrate on locating the centroids and inferring the types of nuclei, by unifying cross-datasets labels to learn a universal model. 

\begin{figure}[!t]
\centering
\includegraphics[width=0.44\textwidth]{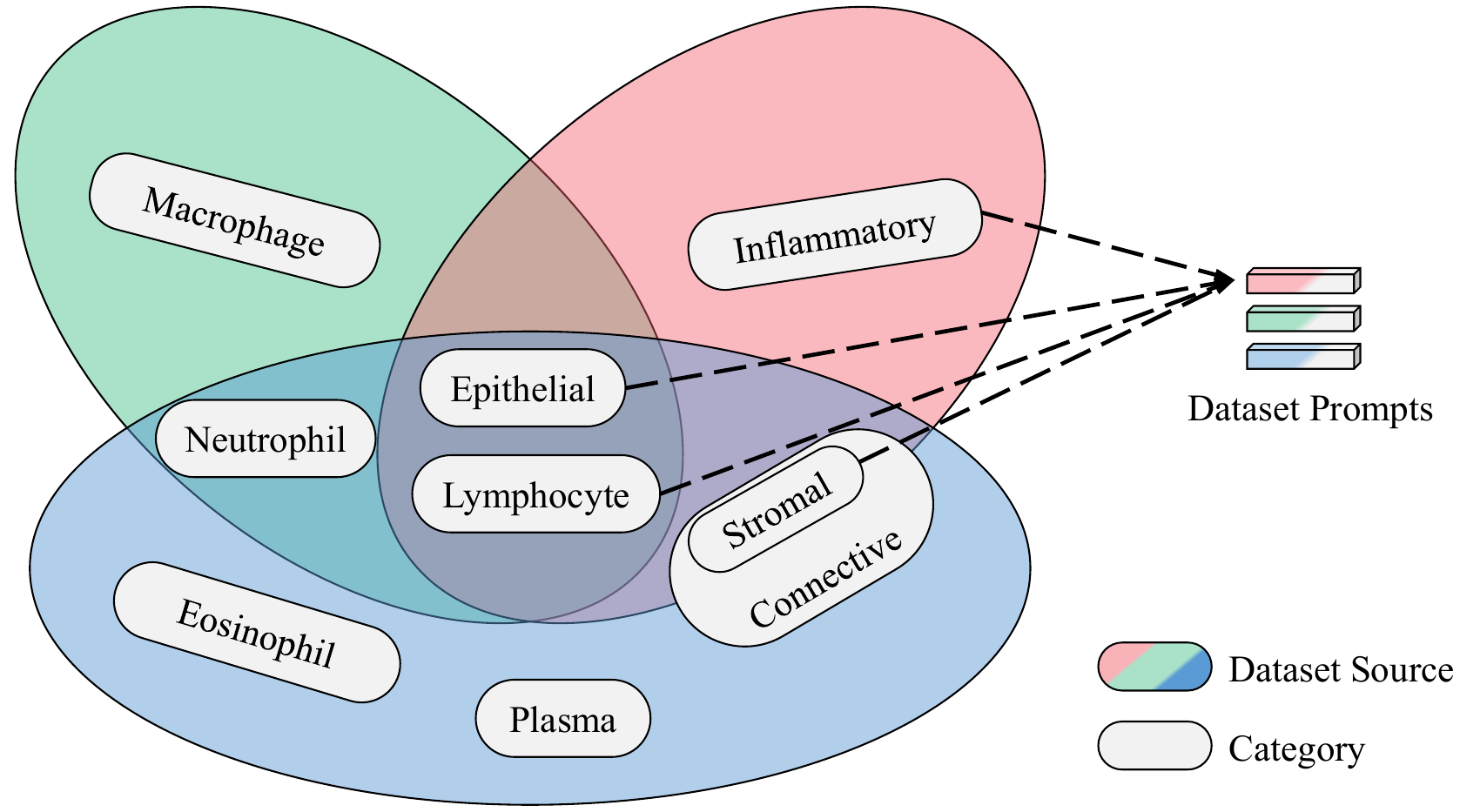} 
\caption{
The illustration of universal multi-dataset cell nucleus classification. The lack of a unified standard for annotating nucleus types hinders the efficient utilization of data and labels. For example, The Lizard dataset (Blue) has three overlapping classes (Neu., Epi., Lym.) with the MoNuSAC dataset (Green). Besides, there exists an inclusion relationship between some categories, such as connective and stromal cells. Our approach utilizes multiple datasets and their associated labels as prompts for training a unified model.}
\label{intro}
\end{figure}

In recent years, numerous pathological datasets~\cite{graham2019hover, graham2021lizard, verma2021monusac2020, ryu2023ocelot} with nucleus annotations have been proposed for deep learning based computational pathology (CPath). However, due to the varying locations of collected organs and the diversity of diseases, the researchers adopt different annotating protocols for these datasets. As Figure~\ref{intro} shows, the annotations from different benchmarks may partially overlap but are not exactly the same. Thus, most existing methods are limited to be trained and evaluated on each single dataset. 
Furthermore, the scale, the diversity and the numbers of samples in pathology benchmarks vary considerably. The labels of some datasets are insufficient for learning a robust model. Consequently, most of previous cell recognition approaches~\cite{abousamra2021multi, doan2022sonnet, huang2022mask, amgad2022nucls} 
fail to adapt to or exploit the common knowledge from multiple data sources.

To overcome the discrepancies among datasets, some recent works~\cite{ tellez2020extending, mormont2020multi, gamper2021multiple, graham2023one} explore more general approaches from a multi-task perspective, which could acquire ample pathological representations to extend to other benchmarks via transfer learning. However, these approaches require an additional training phase to tune a model for each single task or data source, which leads to increasing computational costs and extra training time.  
Some works~\cite{qin2022multi, zhang2023merging} consider merging the labels from distinct sources, employing label semantics to fully utilize all annotated datasets. However, these methods are confined to classifying image patches of cell nuclei, without achieving end-to-end detection and recognition. 

To harvest shared knowledge and prevent redundant training for correlated tasks, we propose an end-to-end framework for cell nucleus detection and classification, and train the single proposed model with multiple datasets concurrently. We resort to prompt learning to represent label semantics and dataset attributes, which enables jointly training a model with multiple data sources and adapting features to various cell types. Specifically, we adopt a DETR based architecture that directly predicts cell locations and categories based on candidate queries, and introduce the contrastive denoising approach to expedite convergence. To fully utilize data from various sources, we employ multiple prediction heads at the end of the proposed model, and the parameters of the model except the heads are shared by different nucleus classification tasks. To well adapt to the categorical semantics from varied sources, we devise a new dynamic prompt module. The module learns the embeddings of category prompts to construct a category memory bank, which is further utilized to yield semantic-aware dataset prompts. After that, image features are enhanced by the dataset prompts in a targeted manner, and become more adaptable to the corresponding tasks. Therefore, we propose an end-to-end model capable of making predictions on multiple data sources. With the benefit of cross-dataset samples, our approach can achieve superior results in cell detection and classification. 
 In short, our major contributions are summarized in three folds:
 \begin{itemize}
    \item We propose UniCell, a novel end-to-end universal nucleus recognition framework, which can learn a single model from multiple datasets concurrently without additional parameters or time for transfer.
    \item We develop a new Dynamic Prompt Module integrated with UniCell. The module can adapt the representation learning to the categorical semantics from different sources using dataset and label prompts. 
     \item We validate UniCell through extensive experiments on four public benchmarks. The results show that our method significantly outperforms the state-of-the-arts.
 \end{itemize}

\begin{figure*}[t]
\centering
\includegraphics[width=0.85\textwidth]{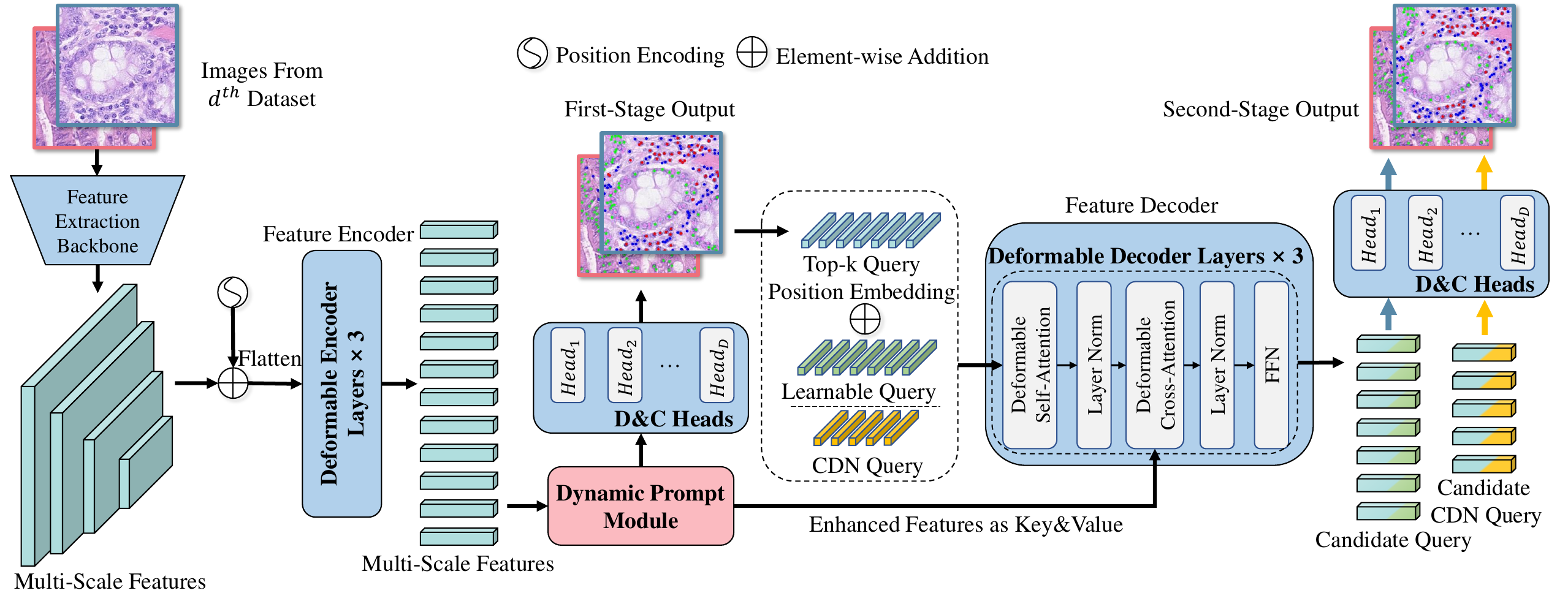} 
\caption{The framework of our proposed UniCell. The multi-scale feature and Dataset ID are inputted into the Dynamic Prompt Module for the specific dataset prompt. Note that both the deep supervision and query refinement in the decoder layer are omitted for readability.}
\label{overall_framework}
\end{figure*}

\section{Related Work}
\paragraph{Cell Nucleus Classification.}
Most traditional solutions to nucleus recognition are based on the Watershed algorithm and handcrafted features~\cite{malpica1997applying, plissiti2010automated, xu2016automatic}, which lack sufficient accuracy and generalization. 
Recently, deep-learning based approaches have achieved notable success. 
Some methods~\cite{abousamra2021multi, wang2022global, sugimoto2022multi} utilize a UNet~\cite{ronneberger2015u} structure to detect cells via pixel-wise binary classification. When given the annotation of nuclei segmentation masks, some methods~\cite{graham2019hover, doan2022sonnet, chen2023cpp} distinguish nuclei by learning to infer the label of each pixel. 
Despite of impressive results, these methods are usually trained independently on a single dataset. 
%
To exploit multiple datasets, most methods~\cite{mormont2020multi, wang2021transpath, yang2022cs,huang2023prompt,lou2023structure} resort to pre-training or transfer learning, which requires additional training costs.
\citet{graham2023one} employs multiple task-specific decoders to merge multi-task data in the model for concurrent training, enabling joint prediction with a single network. Besides, some works~\cite{qin2022multi, zhang2023merging} focus on unifying the labels. \citet{zhang2023merging} formulate the nucleus classification task as a multi-label problem with missing labels. These methods are limited to classification tasks and require multi-stage training. Differently, we build an end-to-end detection and classification model trained with multiple datasets in one stage.

\paragraph{Prompt Learning.}
Prompt learning originally refers to prepending language instruction to the input text allowing the model to better understand the task. Some works~\cite{zhou2022learning, zhou2022conditional, khattak2023maple} input the prompts to the language branch of a Visual-Language model, extracting useful information from the knowledge base for downstream tasks. 
\citet{zhou2022learning} models a prompt's context words with learnable vectors. \citet{jia2022visual} introduce a small number of trainable parameters for the visual model to improve the transfer effect. For universal training, \citet{jain2023oneformer} utilizes the task type as a prompt to distinguish the semantic segmentation and instance segmentation tasks.
Different from existing prompt learning methods, we utilize the fusion of prompt sentences to differentiate feature information and mitigate interference across various datasets.

\section{Methodology}

\subsection{Overall Framework}
The proposed method employs a DETR-like~\cite{carion2020end} structure, which directly predicts of the centroid positions and categories of nuclei without complicated post-refinements. As shown in Figure~\ref{overall_framework}, the proposed UniCell is an end-to-end architecture that has a feature extraction backbone, a feature encoder, a dynamic prompt module, a feature decoder, and detection-classification heads (D\&C Heads).

Assume that we need to solve the nuclei identification task for $D$ different dataset sources whose sets of cell labels are not exactly the same. The $d^{th}$ dataset has $n_d$ samples, and is denoted as a set of training samples, $S_d=\{(x_i^d, y_i^d)\}_{i=1}^{n_d}$. Given a pathological image $x_i^d\in \mathbb{R}^{H\times W\times 3}$ from the $d^{th}$ dataset, we utilize the image $x_i^d$, the ground-truth $y_i^d$ and the corresponding dataset ID $d$ as inputs. 
First, we use the feature extraction backbone Swin Transformer~\cite{liu2021swin} to extract multi-scale features from the input image. Then the extracted features are fed into the feature encoder with their corresponding positional encodings. 
The feature encoder has three deformable self-attention~\cite{zhu2020deformable} layers. After the encoding, the flattened multi-scale features are input to the proposed dynamic prompt module for further feature enhancement. Subsequently, the enhanced representations are processed by the first-stage D\&C Heads for predicting cell positions and categorical probabilities. The D\&C Heads contain $D$ different prediction heads for the $D$ datasets. Each head consists of two parallel fully connected (FC) layers for detection and classification. 

For the $d^{th}$ dataset containing $t_d$ cell types, its D\&C Head outputs the coordinate and a $t_d$-element vector as the categorical confidence of each nuclues. The positions of the top $k$ cell nuclei with the highest confidence are utilized to initialize the query position embedding that describes the locations of anchor centroids and reference points. The query position embeddings are merged with $k$ learnable content queries. In the feature decoder, we compute deformable cross-attention between the $k$ merged queries and the enhanced features to generate $k$ candidate queries. 

\begin{figure*}[t]
\centering
\includegraphics[width=0.9\textwidth]{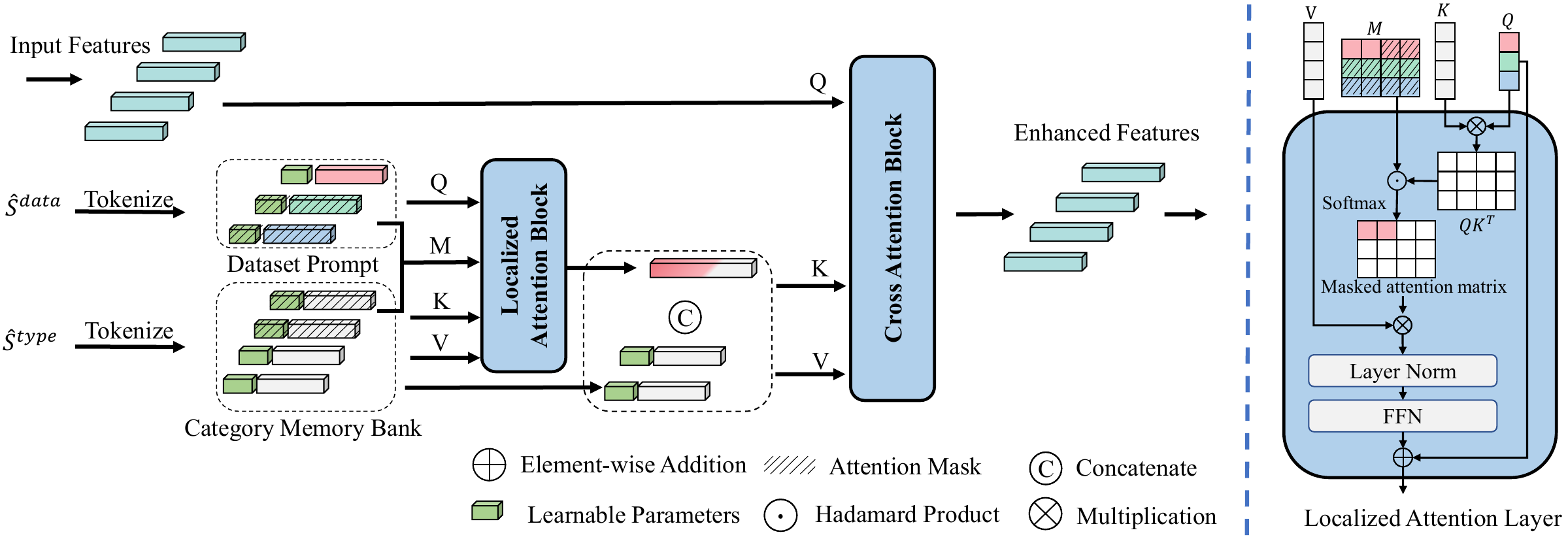} 
\caption{The proposed Dynamic Prompt Module. Both dataset prompt and category memory bank are tokenized and embedded from priori textual sequences. We update the dataset prompts with the embeddings in the category memory bank, and adopt the updated prompts to enhance input representations.}
\label{prompt_module}
\end{figure*}

\paragraph{Contrastive DeNoising Query.} 
Inspired by DINO~\cite{zhang2022dino}, we adopt the Contrastive DeNoising (CDN) approach to avoid duplicate predictions and accelerate training convergence. Specifically, the original CDN method embeds ground truth (GT) labels and bounding boxes with noise to generate positive and negative noise queries, namely CDN queries. The scale of noise is controlled by two hyperparameters $\lambda_1$ and $\lambda_2$ ($\lambda_1 < \lambda_2$). During the training stage, both the CDN queries and the learnable content ones are utilized. During the inference, only the learnable content queries are deployed. Similar to the learnable content queries, we calculate cross-attention between the CDN queries and enhanced features to generate candidate CDN queries, reconstructing the GT independently.

Since we detect and classify nuclei via predicting their centroids, the pathology datasets only have the annotations of centroid locations in our setting. Thus, different from the original CDN, we only apply location-shifting noise to the coordinates of GT centroids. Formally, we denote the GTs as $y_i^d=\{(u_j, v_j), c_j\}_{j=1}^{n_c}$ with $n_c$ annotations. For each annotation, we randomly generate a pair of positive position offset ($|\Delta u_{p}|, |\Delta v_{p}|) \in (0, \lambda_1)$ and a pair of negative one ($|\Delta u_{n}|, |\Delta v_{n}| ) \in (\lambda_1, \lambda_2)$ as noise. The noisy annotation $y_{i\_\{p,n\}}^{d\_{noise}}$ can be formulated as: 
\begin{equation}
    y_{i\_\{p,n\}}^{d\_{noise}} = \{(u_j+\Delta u_{\{p, n\}}, v_j+\Delta v_{\{p, n\}}), RF(c_j, \gamma)\}_{j=1}^{n_c}, 
\end{equation}
where $\textit{RF}(\cdot, \gamma)$ denotes randomly switching the GT labels to others with a probability of $\gamma$. The CDN queries are generated by embedding the locations and categories of $y_{i\_\{p,n\}}^{d\_{noise}}$.

As shown in Figure~\ref{overall_framework}, after the decoder produces the candidate queries and candidate CDN queries, they are sent into the $d^{th}$ head of the second-stage prediction heads, which are similar to the first-stage ones. An independent prediction head is deployed for each dataset. Each head uses two FC layers for detection and classification. We denote the results of candidate queries output by the head as  $\overline{y}_i^d=\{(\overline{u}_j, \overline{v}_j), \overline{c}_j\}^{n_q}_{j=1}$, where $n_q$ is the number of candidate queries. The predictions of positive and negative candidate CDN queries are denoted as $\hat{y}_p^d=\{(\hat{u}_p, \hat{v}_p), \hat{c}_p\}^{n_c}_{p=1}$ and $\hat{y}_n^d=\{(\hat{u}_n, \hat{v}_n), \hat{c}_n\}^{n_c}_{n=1}$, respectively. We adopt Hungarian loss~\cite{kuhn1955hungarian,carion2020end} $H(\cdot)$, L1 loss $L_1(\cdot)$, and Focal loss~\cite{lin2017focal} $\textit{Focal}(\cdot)$ to match the predictions and GTs. The loss function for the $d^{th}$ dataset is formulated as: 
\begin{align}
         \mathcal{L}^d(y_i^d, \overline{y}_i^d, \hat{y}_{\{p,n\}}^d) = & H(y_i^d, \overline{y}_i^d) + L_1(y_i^d, \hat{y}_p^d) + \nonumber \\
         & \textit{Focal}(y_i^d, \hat{y}_p^d) + \textit{Focal}(\widetilde{y}_i^d, \hat{y}_n^d)
\end{align}
where $\widetilde{y}_i^d$ denote empty objects and are set to $t_d\!+\!1$, $L_1(\cdot)$ calculates the Euclidean distance between coordinates and $\textit{Focal}(\cdot)$ measures the category difference.  The overall loss for all the datasets is  $\sum^D_{d=1}\omega_d\mathcal{L}^d(y_i^d, \overline{y}_i^d, \hat{y}_{\{p,n\}}^d),$
where $\omega_d$ denotes a balanced factor for the $d^{th}$ dataset. Additionally, each feature decoder layer produces the offsets to refine the centroid coordinates and reference points of the previous layer or the decoder input. These outputs from intermediate decoder layers are sent to independent prediction heads from deep supervision.

\subsection{Dynamic Prompt Module}
As shown in Figure~\ref{prompt_module}, we propose a Dynamic Prompt Module (DPM), which dynamically adapts the intermediate representations to different dataset sources, via leveraging the dataset name and its label properties as learnable prompts. The DPM takes a sequence of features as input, and outputs the prompt-adapted representations. 
Assume that the $D$ datasets we use have $C$ different cell types. The set of names of these datasets is denoted as: 
$S^{\textit{data}} = \{ \textit{Dataset\_i} \}_{i=1}^{D}$, where \textit{Dataset\_i} is the name of the $i^{th}$ dataset. To compute the prompt embedding of a dataset, we construct a textual sequence (sentence) for each dataset, by adopting learnable unified tokens as Context Optimization (CoOp)~\cite{zhou2022learning} before the dataset name. The textual sequences of all the $D$ datasets are represented as: 
\begin{equation}
    \hat{S}^{\textit{data}} = \{[V_1^{\textit{data}}][V_2^{\textit{data}}] \cdots [V_{T}^{\textit{data}}] [\textit{Dataset\_i}]\}_{i=1}^{D},
\end{equation}
where $\hat{S}_i^{\textit{data}}$ denotes the sentence of the $i^{th}$ dataset and $T$ is the number of tokens. $V_1^{\textit{data}}, V_2^{\textit{data}}, \cdots , V_{T}^{\textit{data}}$ are the learnable tokens, and are shared by the $D$ datasets. `$[*][*]\cdots$' means connecting the tokens into a textual sequence or a sentence.  
We convert each sentence into a lower-cased byte pair encoding (BPE)~\cite{sennrich2015neural} representation with a vocabulary size of 49,408. Each text sequence is encompassed with the [SOS] and [EOS] tokens, and is capped at a fixed length $L^S$ set as the CoOp method~\cite{zhou2022learning}. 
The sequence of size $L^S$ is tokenized and embedded into a 256-dimensional vector. 
All the dataset representations form a $D\!\times\! 256$ matrix, and are also called \textit{Dataset Prompts} as shown in Figure~\ref{prompt_module}.

To compute the common representations for the $C$ categories, we adopt a similar way by building a textual sequence for each category. These sequences are denoted as: 
\begin{equation}
    \hat{S}^{\textit{type}} = \{[V_1^{\textit{type}}][V_2^{\textit{type}}]\cdots [V_{T}^{\textit{type}}][\textit{Category\_i}]\}_{i=1}^{C}, 
\end{equation}
where \textit{Category\_i} is the name of the $i^{th}$ category. The textual sequences of all $C$ categories are also tokenized by the BPE, attached with [SOS] \& [EOS] and capped at the length $L^S$. 
Consider that the category memory bank may need to restore the various patterns of cell types, and should be modeled by multiple vectors. Thus, for each category, its $L^S$ tokens are converted into $L^S$ 256-dimensional embeddings.  Then we obtain a $C\!\times\! L^S\!\times\! 256$ matrix, called \textit{Category Memory Bank}, which is supposed to maintain the common knowledge for identifying cell types.

We first compute the cross-attention between the dataset prompt and the category memory bank, to construct a dataset-specific prompt embedding space. Since a particular dataset may not contain all categories, performing correlation calculations on absent categories would be illogical. Therefore, we employ an attention mask for localized attention computations.
We use $M = [m_{ij}]_{D\times CL^S}$ to denote the attention mask. 
If the category of $j^{th}$ prompt embedding in the memory bank is invisible to the $i^{th}$ dataset, $m_{ij} = -\infty$. Otherwise, $m_{ij} = 0$.
The attention mask is devised as:
\begin{equation}
    m_{ij} = \left\{
        \begin{aligned}
            & 0, \quad if\ i = d \And \left\lceil j/L^S \right\rceil \in I_d \\
            & -\infty, \quad otherwise
        \end{aligned}
    \right.,
\end{equation}
where $I_d$ is the set of category indices belonging to the $d^{th}$ dataset. $\left\lceil j/L^S \right\rceil$ returns the category index of the $j^{th}$ embedding in the memory bank. 

\begin{figure*}[!thbp]
    \centering
    \includegraphics[width=0.8\textwidth]{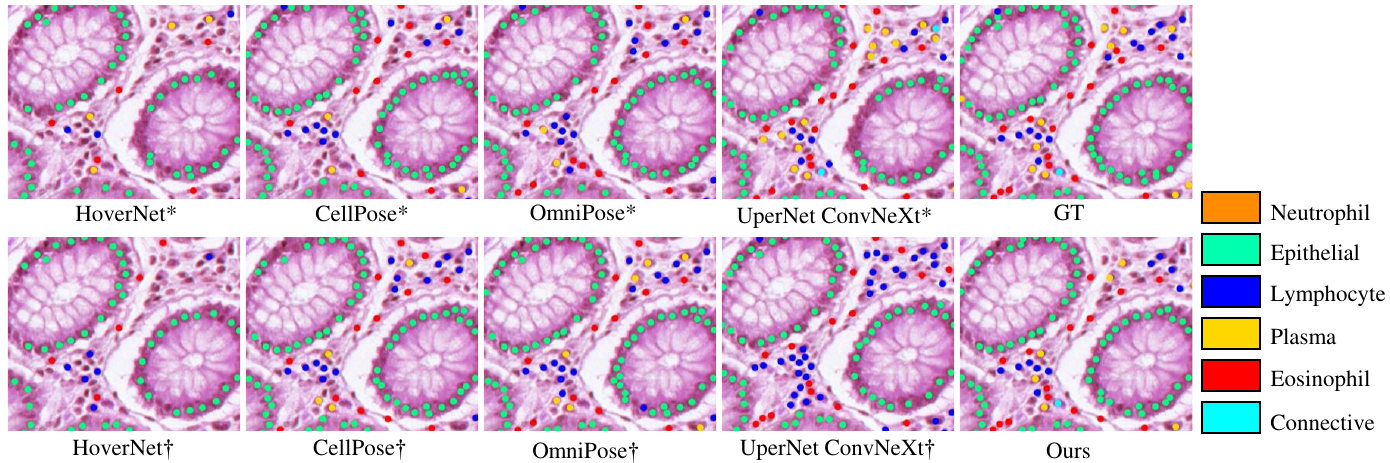}
    \caption{Qualitative comparison on the Lizard dataset. Five types of cells are marked with dilated nucleus centroids in five different colors. As the results show, the category distribution of our method is the closest to that of the ground truths.}
    \label{visual}
\end{figure*}

To embed categorical semantics into the dataset prompts, we set the dataset embeddings to Query $Q$ and the category memory bank to Key $K$ \& Value $V$, respectively. Then we perform the cross-attention operator by sending $Q$, $K$ and $V$ into a localized attention block, 
which contains $L$ localized attention layer. Each layer is formulated as:
\begin{equation}
    Q_{l+1} = \textit{FFN}(\textit{LN}(\textit{CA}(Q_l,K_l,V_l, M)))+Q_l,
\end{equation}
where $l$ is the layer index, $\textit{LN}(\cdot)$ denotes the layer normalization~\cite{ba2016layer}, $\textit{FFN}(\cdot)$ denotes a feed forward network. $\textit{CA}(\cdot)$ denotes a variant of cross-attention guided by an attention mask $M$, and is formulated as:
\begin{equation}
    \textit{CA}(Q_l, K_l, V_l, M) = softmax(M + Q_lK_l^T)V_l,
\end{equation}
where $\odot$ is Hadamard Product. The shape of $Q_l$ / $K_l$ / $V_l$ is $D\times 256$ / $CL^S\times 256$ / $CL^S\times 256$. The localized attention block outputs the updated dataset prompt that is then concatenated with its visible category embeddings in the memory bank, as shown in Fig. \ref{prompt_module}. The concatenated variables serve as Key and Value to be input into a cross attention block as Key and Value to refine the input features of DPM. The cross attention block has $L$ cross attention layers and the input features is set to the Query for attention. 

\noindent\textbf{Feature-Enhancing or Query-Enhancing Strategy.}
In our method, we consider two enhancement strategies: the Feature-Enhancing Strategy and the Query-Enhancing Strategy. The Feature-Enhancing Strategy, depicted in Fig.~\ref{prompt_module}, takes the multi-scale features as the DPM input and updates the features at a pixel level. This allows the model to distinguish fine-grained details across the entire image, thereby facilitating the subsequent identification of cells. 
The query-enhancing strategy takes the learnable content queries as the DPM input, sends the output into the feature decoder. The strategy ensures that each content query is aware of the categorical semantics within the current dataset. Note that when setting the learnable content queries as the DPM input, the unenhanced multi-scale features are directly fed into the feature decoder as keys and values.

\section{Experiments}
\subsection{Implementation Details}
\textbf{Dataset.} We conduct experiments on four datasets: \textbf{CoNSeP}~\cite{graham2019hover}  consists of 41 H\&E stained image tiles from 16 colorectal adenocarcinoma whole-slide images (WSIs). Following the work~\cite{abousamra2021multi}, we resize the images in CoNSeP to 20$\times$ and merge the labels into three types of cells: inflammatory, epithelial, and stromal. 
\textbf{MoNuSAC}~\cite{verma2021monusac2020} contains over 46,000 nuclei from 37 hospitals and 71 patients, sourced from four organs and categorized into four nucleus types: epithelial cells, lymphocytes, neutrophils, and macrophages.
\textbf{Lizard}~\cite{graham2021lizard} is collected from six datasets sources, and has nearly half a million labeled nuclei in H\&E stained colon tissue. These nuclei are categorized into six types of cells: epithelial cells, connective tissue cells, lymphocytes, plasma cells, neutrophils and eosinophils.
To ensure the independence among data sources, we remove the images belonging to the CoNSeP subset from the Lizard dataset.
\textbf{OCELOT}~\cite{ryu2023ocelot} consists of 400 images from six organs with the annotations of nuclei and tissues. The categories of OCELOT include background cells and tumor cells. For OCELOT, we only utilize its nuclues annotations and split the dataset into training and testing sets with a ratio of 7:3. For other datasets, we use their default/official data split. During the training, we combine all training images from four datasets and randomly sample images for model training. During the inference, we evaluate a model on the test set of each benchmark.

\begin{table*}[!t]
\centering
\begin{tabular}{llllllllllll}
    \Xhline{1pt}
    \multirow{2}*{Methods} & \multicolumn{2}{c}{CoNSeP} & &\multicolumn{2}{c}{MoNuSAC} & &\multicolumn{2}{c}{Lizard} & &\multicolumn{2}{c}{OCELOT} \\
     
     \cline{2-3} \cline{5-6} \cline{8-9} \cline{11-12}   
     \specialrule{0em}{2pt}{1pt}
    ~                                           & $F_d$ & $\overline{F_c}$ & & $F_d$ & $\overline{F_c}$ & & $F_d$ & $\overline{F_c}$ & & $F_d$ & $\overline{F_c}$   \\ 
    \Xhline{1pt}
    HoverNet~\cite{graham2019hover}*              & 0.621 & 0.503 & & 0.822 & 0.641 & & 0.729 & 0.430 & & \underline{0.738} & 0.501 \\
    HoverNet~\cite{graham2019hover}$\dagger$      & 0.640 & 0.417 & & 0.802 & 0.341 & & 0.747 & 0.305 & & 0.723 & 0.482 \\
    Cellpose~\cite{stringer2021cellpose}*  & 0.625 & 0.492 & & \underline{0.828} & 0.684 & & 0.783 & 0.450 & & 0.715 & 0.502 \\
    Cellpose~\cite{stringer2021cellpose}$\dagger$ & 0.604 & 0.418 & & 0.826 & 0.650 & & 0.785 & 0.447 & & 0.731 & 0.518 \\
    Omnipose~\cite{cutler2022omnipose}* & 0.678 & 0.528 & & 0.784 & 0.642 & & 0.788 & 0.452 & & 0.716 & 0.509 \\
    Omnipose~\cite{cutler2022omnipose}$\dagger$ & 0.625 & 0.439 & & 0.788 & 0.635 & & \underline{0.799} & 0.461 & & 0.720 & 0.515 \\
    UperNet ConvNeXt~\cite{liu2022convnet}* & \underline{0.715} &  \underline{0.595} & &  0.801 & \underline{0.712} & & 0.764 & \underline{0.474} & & 0.692 & 0.509 \\
    UperNet ConvNeXt~\cite{liu2022convnet}$\dagger$ & 0.652 & 0.418 & & 0.784 & 0.708 & & 0.458 & 0.266 & & 0.732 & \underline{0.540} \\
    \midrule
    UniCell (Ours)& \textbf{0.762} & \textbf{0.679} & & \textbf{0.847} & \textbf{0.752} & & \textbf{0.813} & \textbf{0.596} & & \textbf{0.751} & \textbf{0.596} \\
    \Xhline{1pt}
\end{tabular}
\caption{Comparisons with existing methods on 4 datasets, CoNSeP, MoNuSAC, Lizard and OCELOT. The previous methods are trained in two manners. * denotes the results of training a model on each dataset separately, while $\dagger$ represents the results of training a model with all 4 datasets. For each column, the best method is in bold type and the second best method is underlined.}
\label{SOTA}
\end{table*}

\begin{table*}[!t]
\centering
\scalebox{0.97}{
\begin{tabular}{lllllllllllllll}
    \Xhline{1pt}
    \multirow{2}*{UniCell}  & \multicolumn{2}{c}{CoNSeP} & &\multicolumn{2}{c}{MoNuSAC} & &\multicolumn{2}{c}{Lizard} & &\multicolumn{2}{c}{OCELOT} & & \multicolumn{2}{c}{Average} \\
     \cline{2-3} \cline{5-6} \cline{8-9} \cline{11-12}   \cline{14-15}
     \specialrule{0em}{2pt}{1pt}
    ~       & $F_d$ & $\overline{F_c}$ & & $F_d$ & $\overline{F_c}$ & & $F_d$ & $\overline{F_c}$ & & $F_d$ & $\overline{F_c}$ & & $F_d^{avg}$ & $\overline{F_c}^{avg}$   \\ 
    \Xhline{1pt}
    w/o DPM & 0.734 & 0.650 & & 0.832 & 0.751 & & 0.806 & 0.556 & & 0.728 & 0.556 && 0.775 & 0.628\\
    w/o Category MB & 0.741 & 0.661 & & 0.843 & 0.748 & & 0.805 & 0.578 & & 0.741 & 0.586 & &  0.783 & 0.643\\
    w/o Dataset Prompt & 0.747 & 0.655 & & 0.842 & 0.744 & & 0.809 & 0.573 & & 0.740 & 0.584 & & 0.785 & 0.639\\
    Query-Enhancing & 0.746 & 0.660 & & 0.836 & 0.745 & & 0.808 & 0.580 & & 0.749 & 0.588 & & 0.785 & 0.643\\
    Feature-Enhancing (Ours)& \textbf{0.762} & \textbf{0.679} & & \textbf{0.847} & \textbf{0.752} & & \textbf{0.813} & \textbf{0.596} & & \textbf{0.751} & \textbf{0.596} & & \textbf{0.793} & \textbf{0.656}\\
    \Xhline{1pt}
\end{tabular}
}

\caption{Ablation study of Dynamic Prompt Module (DPM) on four benchmarks. w/o DPM, Category MB and Dataset Prompt mean the results of removing DPM, category memory bank and dataset prompts from our proposed method, respectively. Query-Enhancing is to deploy DPM to enhance learnable content queries before the feature decoder, while Feature-Enhancing (Ours) is to refine multi-scale features with DPM (Figure~\ref{overall_framework}). 
For each column, the best method is in bold.}
\label{ab:dpm}
\end{table*}

\noindent\textbf{Evaluations.} We utilize the F-score as the evaluation metrics for nucleus detection and classification tasks, following the works~\cite{graham2019hover}. For the detection, we compute the Euclidean distance between each predicted centroid and GT to yield a cost matrix. Then we run the Hungarian algorithm~\cite{kuhn1955hungarian} with the cost matrix to obtain the paired results. The paired results with a distance less than a predefined radius $r$ are correctly detected centroids ($\textit{TP}_d$, $d$ for detection) while the rest of paired results and unpaired predicted centroids are overdetected predicted centroids ($\textit{FP}_d$). The unpaired GT centroids are called misdetected GT ($\textit{FN}_d$). The detection F-score is calculated as: 
$F_d=\frac{2\textit{TP}_d}{2\textit{TP}_d+\textit{FP}_d+\textit{FN}_d}$. We set the radius $r$ as 15 pixels for OCELOT and 6 pixels for the other three datasets, following the works~\cite{ryu2023ocelot,sirinukunwattana2016locality}.

For the classification task with $K$ classes, $\textit{TP}_d$ are split into multiple subsets: correctly classified centroids of Type $k$ $(\textit{TP}^k_c)$, incorrectly classified centroids of Type $k$ ($\textit{FP}^k_c$) and incorrectly classified
centroids of types other than Type $k$ ($\textit{FN}^k_c$). The classification F-score is defined as: $F^k_c = \frac{2\textit{TP}^k_c}{2(\textit{TP}^k_c+\textit{FP}^k_c+\textit{FN}^k_c)+\textit{FP}_d+\textit{FN}_d}$. The average F-score of classification is $\overline{F_c}=\sum_{k=1}^K F_c^k$.

\noindent\textbf{Hyper-parameters setting.}
We adopt Swin-B~\cite{liu2021swin} with ImageNet~\cite{deng2009imagenet} pre-trained weights as the backbone. The number of both feature encoder and decoder layers is set to 3. The number of candidate queries is 900. The learnable tokens number $M$ is 16. The number of Localized Attention Layer $L$ is 3. AdamW optimizer is used to train UniCell with initial learning rates of $1e^{-4}$ and $1e^{-5}$ for backbone and other modules, respectively. 
For data argumentation, we apply the random flip, random crop and multi-scale training with sizes between 600 and 800, and infer images after resizing to 800$\times$800. All models are trained and tested with an NVIDIA A100 (80GB) GPU. The number of training iterations is set to 160k and after training for 160k iterations, we choose the final model for evaluation. 
We use the SAHI~\cite{akyon2022slicing} scheme to slice the images into fixed-size patches as training samples and adopt sliding-window prediction during inference.

\subsection{Comparison with State-of-the-arts}
In Tab. \ref{SOTA}, we compare our proposed method with existing methods HoverNet~\cite{graham2019hover}, Cellpose~\cite{stringer2021cellpose}, Omnipose~\cite{cutler2022omnipose} and UperNet with ConvNeXt~\cite{liu2022convnet} backbone. 
We try to train each of existing methods with all 4 datasets concurrently to build a single model by merging cross-sources labels into 9 categories, which is denoted as $\dagger$. These models use the same training data as ours. We also train each method on each dataset separately, which is denoted by * in Tab. \ref{SOTA}. Our method achieves the highest mean F-score in both detection and classification tasks on all four benchmarks. Fig. \ref{visual} presents a qualitative comparison between the previous approaches and ours. 

In Tab. \ref{SOTA}, our method obtains 4.7\%, 1.9\%, 1.4\% and 1.3\% F-score in detection higher than the second best results on four datasets. 
The results show that our method effectively addresses detection obstacle caused by the various image sizes and nucleus densities across different datasets.
%
Since the categories in each dataset do not entirely overlap, it may cause a more severe class imbalance issue after merging datasets. 
For the classification task, our method outperforms the second best method by 8.4\%, 4\%, 12.2\% and 5.6\% F-scores on four benchmarks, respectively. As the results suggest, our method with the dynamic prompt module effectively leverages the data from various sources to improve the model ability. 

\subsection{Ablation Study}

\noindent\textbf{Effectiveness of the proposed dynamic prompt module (DPM).}
In DPM, we utilize both the dataset prompt and category memory bank to enhance image features. 
To validate the efficacy of DPM, category memory bank (CMB) and dataset prompts, we examin the UniCell model after removing each of the three components, respectively. When removing dataset prompts, the category memory bank serves as keys and values to enhance multi-scale features. 
Comparing our method with `w/o DPM' shows that DPM improves the baseline by 1.8\% in $F_d^{avg}$ and 2.8\% in $\overline{F}_c^{avg}$ on the four datasets. Notably, DPM proves beneficial for the dataset with fewer samples, such as CoNSeP, and the dataset with fewer categories, such as OCELOT. These results show that DPM can well adapt the visual features to different datasets via exploiting their common semantic knowledge.

\begin{figure}[!t]
    \centering
    \subfloat[$\frac{F_d-Method}{F_d-Ours}$]{\includegraphics[width=.48\columnwidth]{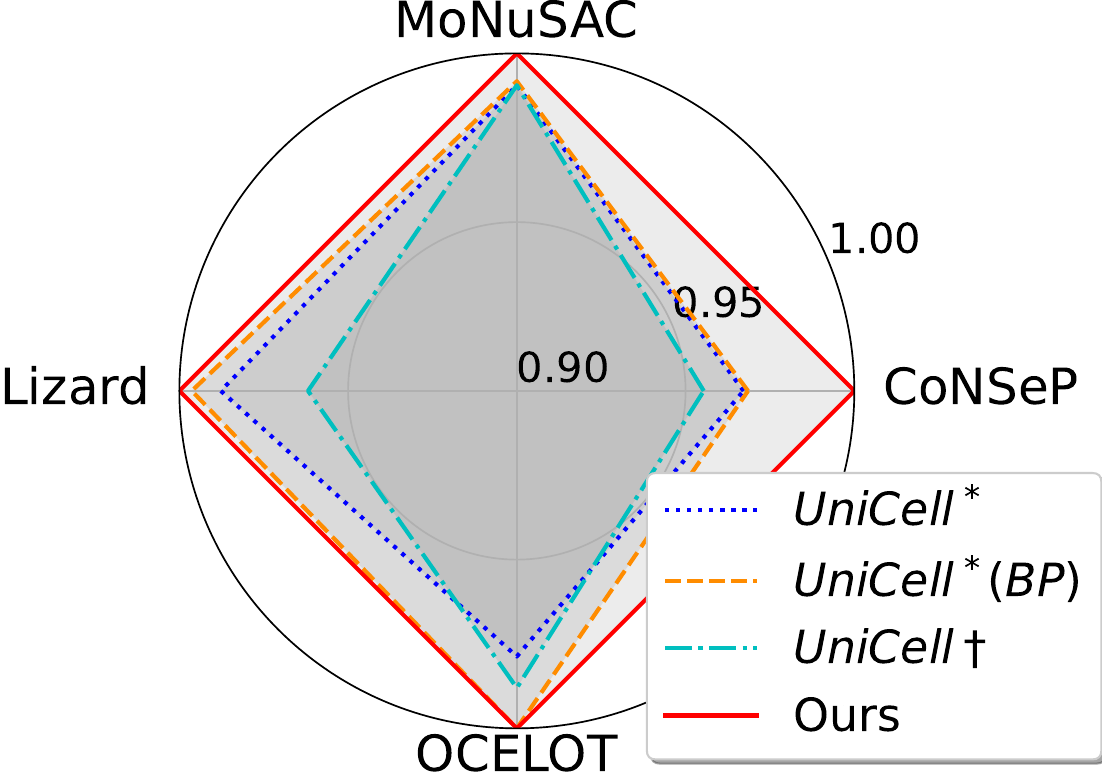}}
    \subfloat[$\frac{\overline{F_c}-Method}{\overline{F_c}-Ours}$]{\includegraphics[width=.48\columnwidth]{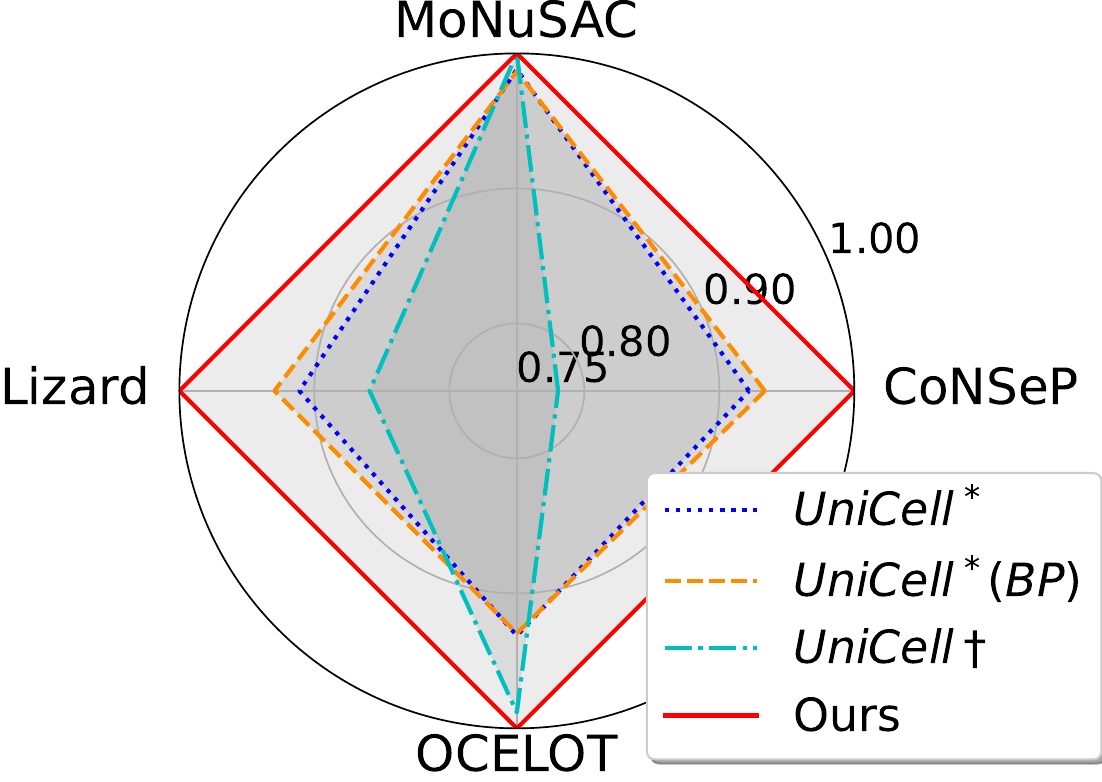}}
    \caption{Different ways of using multiple datasets. `UniCell*' refers that UniCell with one (instead of four) head is trained independently on each dataset after removing DPM. `UniCell*(BP)' denotes training `UniCell*' from the weights pre-trained on a binary detection \& classification task. 
    `UniCell$\dagger$' uses the training data of four sources to train a UniCell* model by merging the categories of all datasets. 
    }
    \label{ab:detcls}
\end{figure}

Besides, we study the proposed two deploying strategies of DPM, query-enhancing and feature-enhancing. The query-enhancing strategy deploys DPM before the feature decoder to refine content queries, which only provides instance-level information from dataset and category prompts.
In contrast, the feature-enhancing strategy enhances multi-scale features at a pixel-level granularity, and the enhanced representations are also utilized to predict instance-level content queries. 
As shown in Tab. \ref{ab:dpm}, both strategies are effective to improve the baseline w/o DPM by 1.5\%-2.8\% $\overline{F}_c^{avg}$. The feature-enhancing strategy is the better, exceeding the other by 1.3\% in $\overline{F}_c^{avg}$. Thus, we adopt the feature-enhancing strategy in our model by default.

\noindent\textbf{Effectiveness of the way to merge.} In Fig. \ref{ab:detcls}, we present the results of exploiting different datasets in various manners. `UniCell*' denotes the UniCell that has one (instead of four) head and is independently trained on each dataset after removing DPM.  
To enable the model to employ the information of all datasets, we pretrain the UniCell* model on a binary detection and classification task using all datasets, and then finetune the model on the multi-label classification tasks using each single dataset, respectively. The resulting model is denoted as `UniCell*(BP)' with BP for `binary-class pretraining'.
UniCell$\dagger$ denotes training a one-head model with all training sets after merging their categories. Fig.~\ref{ab:detcls}(b) shows that neither pre-training nor combined category training effectively improve the baseline of individual training (UniCell*). In contrast, our method manages to exploit the cross-sources knowledge and achieve better performance on all datasets with just one training stage.

\begin{figure}[!t]
    \centering
    \includegraphics[width=0.98 \columnwidth]{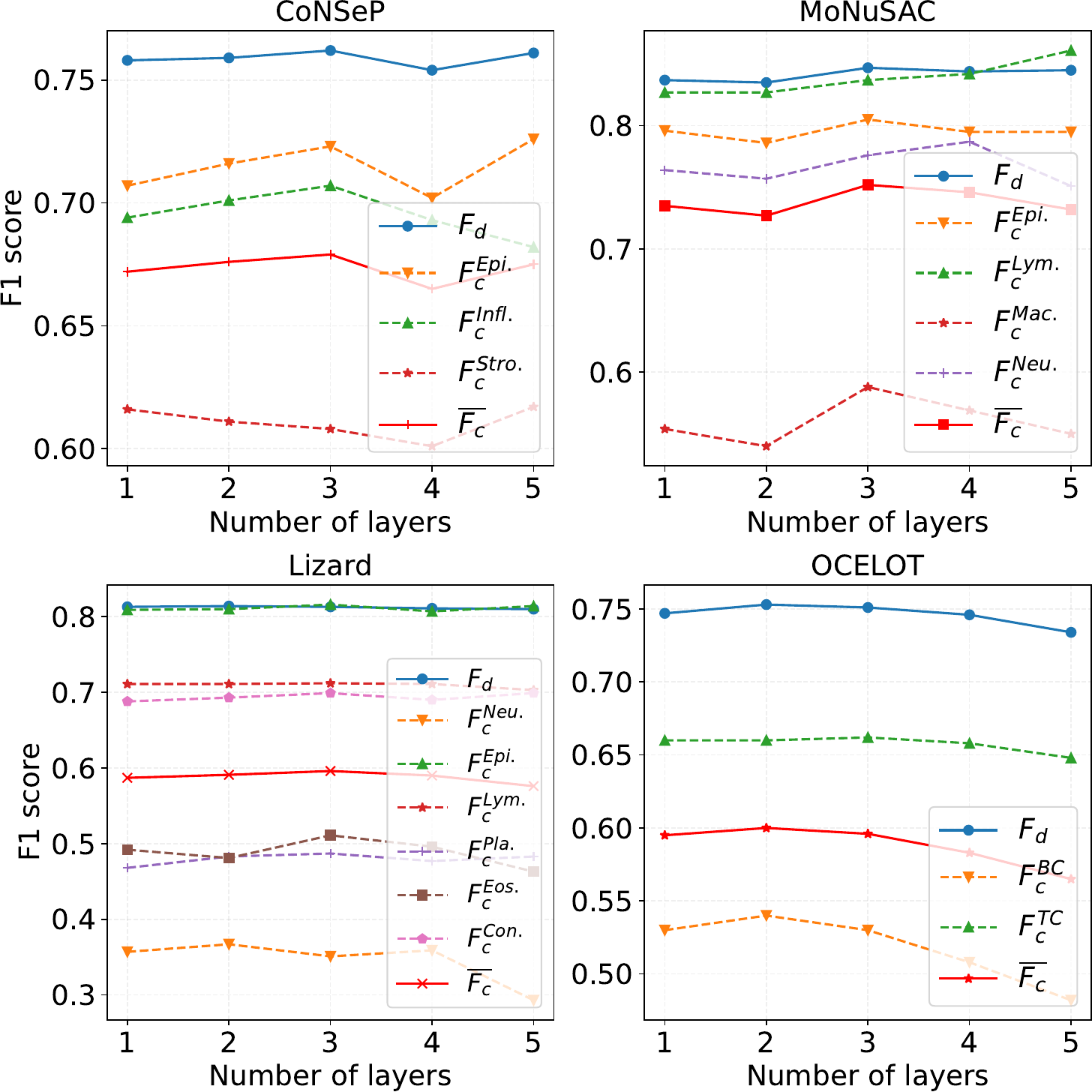}
    \caption{The effects of the number of Localized Attention Layers. $F_c^{Epi.}$, $F_c^{Infl.}$, $F_c^{Stro}$, $F_c^{Lym.}$, $F_c^{Mac.}$, $F_c^{Neu.}$, $F_c^{Pla.}$, $F_c^{Eos.}$, $F_c^{Con.}$, $F_c^{BC}$ and $F_c^{TC}$ are the F-scores of the epithelial, inflammatory, stromal, lymphocytes, macrophages, neutrophils,  plasma, eosinophil, connective tissue cells, background cells and tumor cells, respectively. }
    \label{ab:num_layer}
\end{figure}

\noindent\textbf{Amount of the Localized Attention Layer.} Fig. \ref{ab:num_layer} shows the results of using different numbers of Localized Attention Layers $L$. We provide the classification F-scores for each category in each dataset as well as the detection F-scores. The results show that setting $L$ to 3 performs the best. The model performance does not vary significantly for smaller amounts of $L (L \leq 3)$. The increase of $L$ degrades the performance of a small number of classes, such as neutrophils, macrophages and background cells.

\section{Conclusion}
In this paper, we propose an end-to-end universal cell nucleus recognition architecture (UniCell), which can locate and identify multi-category nuclei of cross-dataset pathological images using a single model. 
We build the overall framework based on DETR that enables the direct detection and classification of nuclei without post-processing, and introduce Contrastive DeNoising to achieve fast and robust convergence. Importantly, to adapt to multiple data sources, we develop a novel dynamic prompt module that employs dataset- and class-specific knowledge as priors to guide the feature extraction. 
We conduct extensive experiments to compare our UniCell with state-of-the-art methods, and validate the effectiveness of the proposed DPM on four challenging benchmarks. With the promising results, we believe that our method has great potential for cell nucleus classification and can bring new insights to the community.

\section{Acknowledgments}
This work was supported in part by the National Natural Science Foundation of China (NO.~62322608, NO.~62102267), in part by the Guangdong Basic and Applied Basic Research Foundation (2023A1515011464), in part by the Shenzhen Science and Technology Program JCYJ20220818103001002), and in part by the Guangdong Provincial Key Laboratory of Big Data Computing, The Chinese University of Hong Kong, Shenzhen.

\bibliography{aaai24}

\end{document}